%% file: main.tex
\documentclass{article} 
\usepackage[dvipsnames]{xcolor}
\usepackage{listings}
\usepackage[ruled,vlined, algo2e]{algorithm2e}
\usepackage{subcaption}

\usepackage{iclr2023_conference,times}

\input{math_commands.tex}

\usepackage{tikz}
\usepackage{pgfplots}
\usepgfplotslibrary{external}
\usetikzlibrary{shapes,external}
\usepackage{hyperref}
\usepackage{url}
\usepackage{graphicx}
\usepackage{caption}
\usepackage{subcaption}
\usepackage{algorithm}
\usepackage{booktabs}
\newtheorem{theorem}{Theorem}[section]
\newtheorem{lemma}[theorem]{Lemma}
\usepackage{wrapfig}
\usepackage{makecell}
\usepackage{multirow}
\usepackage{ifsym}

\usepackage{colortbl}
\usepackage{tipa}
\allowdisplaybreaks

\iclrfinalcopy

\title{PREF: Phasorial Embedding Fields for Compact Neural Representations}

\author{
Binbin Huang$^1 \thanks{Equal Contributions}$ , \ \  Xinhao Yan$^{1*}$, \ \  Anpei Chen$^2$, \  Shenghua Gao$^{1\dag}$  \& Jingyi Yu$^{1\dag}$ \\
$^1$ShanghaiTech University \quad $^2$ETH Z\"urich \\
{$^\dag$Corresponding:\texttt{\{gaoshh,yujingyi\}@shanghaitech.edu.cn}} \\
{\  \footnotesize \url{https://github.com/hbb1/PREF}} \\
}

\begin{document}

\maketitle

\begin{abstract}
We present an efficient frequency-based neural representation termed PREF: a shallow MLP augmented with a phasor volume that covers significant border spectra than previous Fourier feature mapping or Positional Encoding. At the core is our sparse phasor volume that covers a dense spetra. However, computing dense Fourier features is prohibitively expensive. To this end, we develop a tailored and efficient Fourier transform that combines both Fast Fourier transform and local interpolation to accelerate na\"ive Fourier mapping. We also introduce a Parsvel regularizer that stables frequency-based learning. In these ways, Our PREF reduces the costly MLP in the frequency-based representation, thereby significantly closing the efficiency gap between it and other hybrid representations, and improving its interpretability. Comprehensive experiments demonstrate that our PREF is able to capture high-frequency details while remaining compact and robust, including 2D image generalization, 3D signed distance function regression and 5D neural radiance field reconstruction.

\end{abstract}

\input{iclr2023/section/1_introduction}
\input{iclr2023/section/2_related}

\input{iclr2023/section/3_method}
\input{iclr2023/section/4_experiment}
\input{iclr2023/section/6_conclusion}

\bibliography{iclr2023_conference}
\bibliographystyle{iclr2023_conference}
\newpage
\appendix
\input{iclr2023/section/7_appendix}
\end{document}

%% file: math_commands.tex
\usepackage{amsmath,amsfonts,bm}

\def\eqref#1{equation~\ref{#1}}

\def\floor#1{\lfloor #1 \rfloor}
\def\1{\bm{1}}

\def\rphi{{\textnormal{$\phi$}}}

\def\rd{{\textnormal{d}}}
\def\re{{\textnormal{e}}}
\def\rf{{\textnormal{f}}}

\def\ri{{\textnormal{i}}}
\def\rj{{\textnormal{j}}}
\def\rk{{\textnormal{k}}}

\def\rn{{\textnormal{n}}}

\def\ru{{\textnormal{u}}}
\def\rv{{\textnormal{v}}}
\def\rw{{\textnormal{w}}}
\def\rx{{\textnormal{x}}}
\def\ry{{\textnormal{y}}}
\def\rz{{\textnormal{z}}}

\def\rvf{{\mathbf{f}}}

\def\rvu{{\mathbf{i}}}

\def\rvk{{\mathbf{k}}}

\def\rvp{{\mathbf{p}}}

\def\rvu{{\mathbf{u}}}
\def\rvv{{\mathbf{v}}}
\def\rvw{{\mathbf{w}}}
\def\rvx{{\mathbf{x}}}

\def\rmF{{\mathbf{F}}}

\def\rmK{{\mathbf{K}}}

\def\rmP{{\mathbf{P}}}

\DeclareMathAlphabet{\mathsfit}{\encodingdefault}{\sfdefault}{m}{sl}
\SetMathAlphabet{\mathsfit}{bold}{\encodingdefault}{\sfdefault}{bx}{n}



%% file: iclr2023/section/1_introduction.tex
\section{Introduction}
We recently witness considerable advances in implicit neural representations that remain fast and compact while being capable to render high-frequency details \citep{martel2021acorn,sun2021direct,yu2021plenoxels,muller2022instant,chen2022tensorf}.
To trade memory for speed,
they are typified by a hybrid representation:
a shallow but efficient MLP augmented with a luxury data structure like dense grid  \citep{hedman2021baking,sun2021direct}, tree \citep{yu2021plenoxels, NGLOD}, point cloud \citep{xu2022point} or mesh \citep{yang2022neumesh, Munkberg_2022_CVPR}. As usual, they encode the inputs $\rx$, \emph{i.e.}, coordinates, into a high dimensional embedding field $\rvf(\rx)$:
\begin{equation}
    \rvf(\rvx) = \sum_{\ri}^{N_\rvv} \rphi(\rvx-\rvx_\ri) \cdot \rvv_\ri
\end{equation}
where $\rvv_i\in\mathbb{R}^{k}$ is the learnable embedding vector associated with an often fixed and data structure-dependent location $\rx_\ri \in \mathbb{R}^{d}$. $\rphi(\rx)$ is an interpolation kernel, usually local and linear, to achieve an efficient query. 
To recover high frequencies, they require discretizing the spatial location at a very fine level that results in a high memory footprint \citep{sun2021direct}, or joint co-adapt the vertices (by splitting, pruning or subdivision) into a heuristic structure like tree or mesh to capture high-frequency locals \citep{martel2021acorn}. As a result, they may complicate the training process or limit the representation to be task-specific.

\par
By contrast, frequency encoding does not impose any specific data structure to recover high frequencies \citep{mildenhall2020nerf, tancik2020fourfeat}. In fact, it facilitates easy queries and optimizations of specific frequencies of the input embedding field, despite relying on a mathematically simple (inverse) 
Fourier transform:
\begin{equation}
\rvf(\rx) = \sum_{\ri}^{N_k} \re^{j2\pi \rvk_\ri^T\rvx} \cdot \rvp_i 
\label{eq:fourier_transform}
\end{equation}
where $\rvp_i$ is a complex-valued vector, associated with a frequency coordinate $\rk_\ri$. We call $\rvp_i$ as phasor because its entry implies a wave of a certain frequency, phase and magnitude. This process can be implemented equally as matrix production, \emph{i.e.}, 
$f(\rvx) = \rmP [\sin(2\pi\rmK\rvx), \cos(2\pi \rmK\rvx)]$, where $\rmP\in \mathbb{R}^{k\times 2N_k}$ and $\rmK = [\rvk_0,\rvk_1,\cdots, \rvk_{N_k-1}]$. 
Now, Let us consider the distribution of the Fourier basis $\rmK$. If $\rmK$ is uniform-distributed, from the Nyquist-Shannon sampling perspective, a phasor volume should have the same expressiveness (bandwidth) as a uniform spatial volume with the same size,  \emph{i.e.}, $N_\rvv = N_\rvk$. 
In fact, frequency representation turns out to be more compact than such a dense parameterization scheme by directly accessing sparse frequencies. \cite{tancik2020fourfeat} demonstrate  Fourier mapping with a set of sparse sampled Fourier basis outperforms that of uniform-spaced (and therefore a dense grid). One explanation is that natural signals do exhibit some smoothness or correlations, so an MLP augmented with relatively sparse spectra is sufficient.
Despite being compact, frequency representations seem less attractive compared to hybrid representations \citep{sun2021direct, muller2022instant}, mainly because of their inefficiency in Fourier mapping. For example, NeRF's MLPs \citep{mildenhall2020nerf} use only limited spectrum, \emph{e.g.}, $ N_\rk = 24$ as opposite to $N_\rv = 160^3$ used in a dense volume \citep{sun2021direct}. As a result, they should equip a costly larger MLP that incurs longer training and inference time. To allow a better trade-off between memory and speed, it is desirable to scale up spectra and improve the transform efficiency.

\begin{figure}
\vspace{-10pt}
    \centering
    \includegraphics[width=1.0\textwidth]{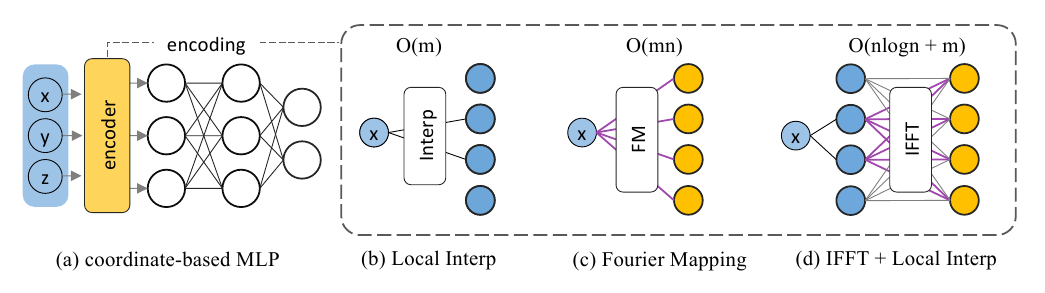}
    \caption{
    A conceptual overview of accelerating frequency-based implicit neural representation. Assume there are $\mathrm{m}$ coordinates in a single iteration and $\rn$ embedding nodes in the encoding layer.
    }
    \label{fig:overview}
    \vspace{-5pt}
\end{figure}

To this end, 
we propose a PhasoRial Embedding Field: a shallow MLP augmented with a phasor volume that covers significant border spectra than previous frequency representations \citep{tancik2020fourfeat}.
To retain compactness and efficiency, we devise a phasor volume where frequencies distribute uniformly along a 2D plane and dilate along a 1D axis. 
Then, we develop a tailored approximated algorithm that combines both Fast Fourier transform (FFT) and local interpolation to accelerate na\"ive Fourier mapping. To make PREF stable, we also propose a Parsvel regularizer tailored for the phasor layer, an analogy to the Total Variation (TV) regularizer in local parameterization schemes. Moreover, the design of PREF improves interpretability by assigning the majority of network capacity to the phasor volume instead of the black-box MLPs. As a result, we can modulate the phasor embedding to edit the neural field, such as level-of-details filtering, which is either inconvenient or even not applicable for the recent rising hybrid encoders. 

\par
To summarize, the contributions of our work include:
(1) A compact and efficient approach called PREF that is more cable of capturing high-frequency details, better than all previous frequency-based neural representations and comparative to existing hybrid representations, such as dense grid and Octree. Extensive experiments, ranging from 2D image generalization, and 3D signed distance function fitting to 5D radiance field reconstruction, illustrate this.
(2) A Parsvel regularizer, tailored for frequency encoding, is proposed as an analogy to other regularizers used in fully connected layers or local interpolation layers.

%% file: iclr2023/section/2_related.tex
\section{Related Work}

Our PREF framework is in line with renewed interest in adopting implicit neural networks to represent continuous signals from low dimensional input. In computer vision and graphics, they include 2D images, 3D surfaces in the form of occupancy fields \citep{mescheder2018occupancy,DVR,oechsle2021unisurf,peng2020convolutional} or signed distance fields (SDFs) \citep{SAL, IGR, wang2021neus, yariv2020multiview, martel2021acorn, NGLOD, yariv2021volume, wang2021spline, park2019deepsdf}, concrete 3D volumes with a density field \citep{ji2017surfacenet,qi2016volumetric}, 4D light fields \citep{levoy1996light,wood2000surface} and 5D plenoptic functions for the radiance fields \citep{mildenhall2020nerf,zhang2020nerf++, liu2020neural, park2021nerfies, yariv2020multiview, chen2022mobilenerf}. 

\noindent \textbf{Frequency encoding.} To learn high frequencies, state-of-the-art implicit neural networks have adopted the Fourier encoding scheme by transforming the coordinates with periodic $\sin(\rx)$ and $\cos(\rx)$ functions or equivalently, under Euler's formula $\re^{\rj\rx}$. 
Under Fourier encoding \citep{tancik2020fourfeat, zhong2021cryodrgn, mildenhall2020nerf, liu2021zero}, feature optimization through MLPs can be mapped to optimizing complex-valued matrices with complex-valued inputs in the linear layer.
Specifically, 
Position Encoding (PE) and Fourier Feature Maps (FFM) both transform spatial coordinates to Fourier basis at earlier input layers whereas SIREN \citep{sitzmann2019siren} and MFN \citep{fathony2020multiplicative} embed the process in the deeper layers by using periodic activation functions. 
In the computer graphics community, frequency representation has used Fourier coefficients to represent opacity \citep{jansen2010fourier} and spherical harmonics (SH) \citep{muller2006spherical} to represent view-dependent effect.
In the context of neural fields for graphics, accessing frequency representations has also led to many other advances. For example, frequency representations allow level-of-detail smoothing and multi-scale anti-aliasing rendering \citep{ lindell2022bacon, barron2021mip}. Like BACON \citep{lindell2022bacon} and FFM \citep{tancik2020fourfeat}, the output of our phasor encoder can be also expressed as a summation of sines. Whereas their frequencies are randomly sampled and the transformation is exactly computed, our frequency support is tailored so that 
allows a fast and approximated Fourier Transform. Fig. \ref{fig:overview} briefly outlines the differences. Beyond neural fields,  Fourier-CPNN \citep{tesfaldet2019fourier} has proposed to learn Fourier coefficients to synthesize the image. However, the coefficients are learned instead of optimized from scratch so that they may incorrectly admit continuity in the spectra, and consequently, hinders the performance.

\textbf{Hybrid representation.} Improving the training and inference efficiency of MLP-based networks has been explored from the embedding perspective with smart data structures. Various schemes \citep{reiser2021kilonerf,hedman2021baking,yu2021plenoctrees,yu2021plenoxels,sun2021direct,muller2022instant,chen2022tensorf} replace the deep MLP architecture with voxel-based representations, to trade memory for speed.  
Early approaches bake an MLP to an Octree along with a kilo sub-NeRF or 3D texture atlas for real-time rendering \citep{reiser2021kilonerf,hedman2021baking}. More smart and compact data structure includes point cloud \citep{xu2022point} and triangle mesh \citep{Munkberg_2022_CVPR, yang2022neumesh}. These approaches, however, rely on a pre-trained or half-trained MLP as prior and therefore still incur long per-scene training and complicate the training process. 
Plenoxel \citep{yu2021plenoxels} and DVGO \citep{sun2021direct} directly optimize the density values on discretized voxels and employ an auxiliary shading function represented by either spherical harmonics (SH) or a shallow MLP network to account for view dependency. They achieve orders of magnitude acceleration over the original NeRF on training but incur a very large memory footprint by storing per-voxel features. Besides, over parametrization can easily lead to noisy density estimation and subsequently inaccurate surface estimations and rendering. 
Seminal works have attempted to store and query voxel features in an easier and much more compact way.
Instant-NGP \citep{muller2022instant} stores and looks up features with a hash table to achieve unprecedentedly fast training and very visual results. EG3D \citep{Chan2022} queries feature from a compact tri-plane representation. In a similar vein, TensoRF \citep{chen2020tensor} employs highly efficient tensor decomposition via vertical projections. Whereas these methods differ in how they look up the very local substitute, our method computes feature substitute that relates to the global parameter space through an approximated Fourier transform. The difference is that the frequency of the field is now directly related to the input phasor field, and the optimization is not local anymore so that PREF remains robust to recover high frequency details, reducing the risk of overfitting to certain locals.

%% file: iclr2023/section/3_method.tex
\begin{figure}
\vspace{-17pt}
\centering
\includegraphics[width=0.94\textwidth]{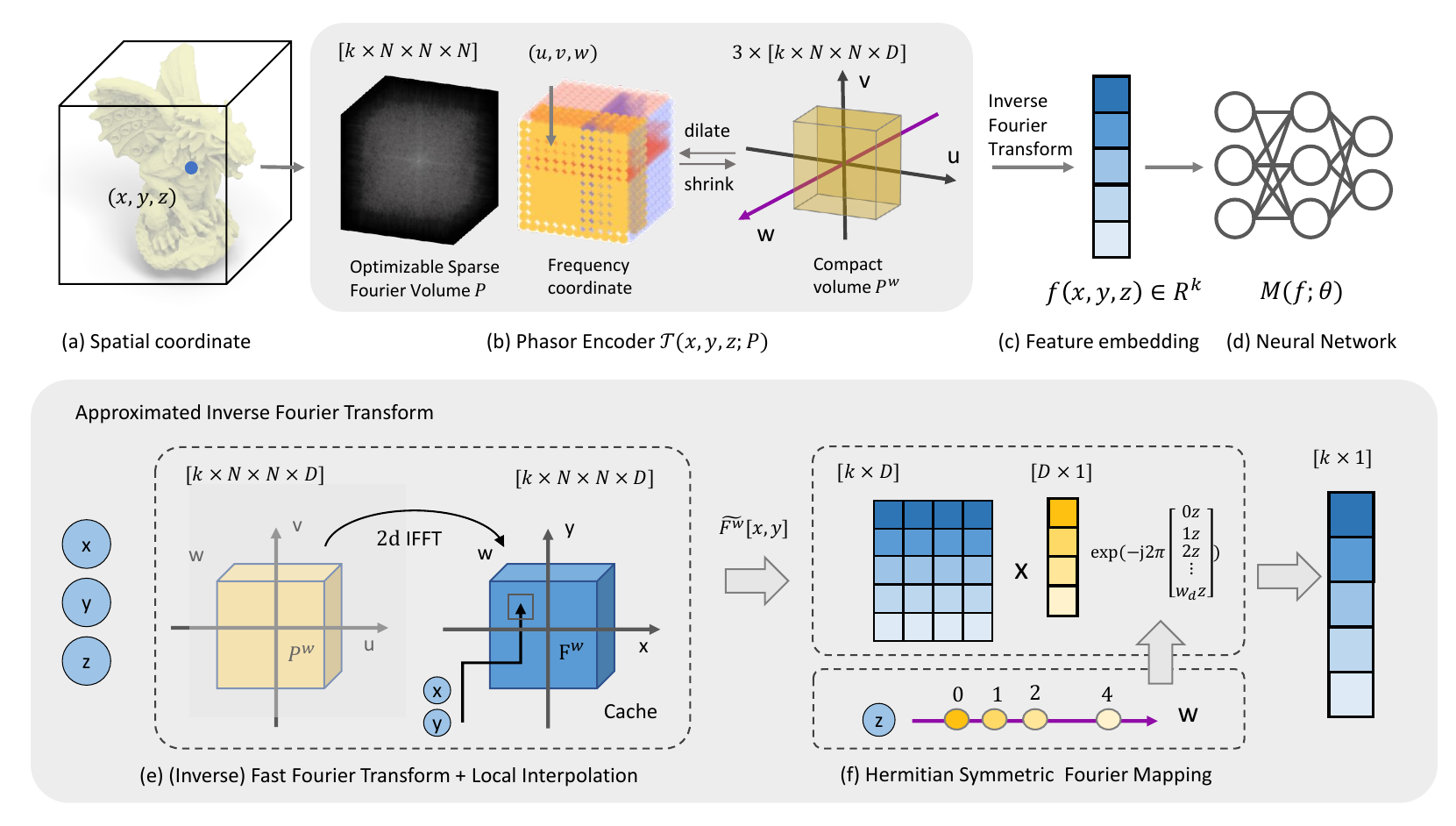}
\caption{Pipeline of PREF. We model the underlying sparse frequency representation of the scene   (b) and augment it with neural network (d) to achieve full spectrum recovery. The inverse Fourier transform is approximated by two consecutive sub-procedures (e) and (f).
} 
\vspace{-8pt}
\label{fig:PREF}
\end{figure}

\section{Phasorial Embedding fields}
PREF consists of a dense frequency encoder and a shallow MLP. We illustrate our design with a 3D signal for example.
We denote the phasor volume $\rmP$ as a 3D complex-valued feature volume with embedding length $\rk$, and denote the encoder as $\mathcal{T}$ thus a local high dimensional embedding is obtained by $\rvf(\rx) = \mathcal{T}(\rx;\rmP) \in \mathbb{R}^{\rk}$ with $\rvx = (\rx,\ry,\rz)$. Although other time-frequency transforms exist, this paper focus on the simple Fourier transforms so as to align with previous frequency encoders \citep{mildenhall2020nerf, tancik2020fourfeat}. 
We first introduce the design of the phasor volume $\rmP$ (Sec. \ref{sec:volume}), and along the way, describe its tailored approximated Fourier transform $\mathcal{T}$ (Sec. \ref{sec:aft}) and regularizer  $\mathcal{L}_\mathrm{reg}$ (Sec. \ref{sec:reg}). We provide an overall visual depiction in Fig. \ref{fig:PREF}. We also include the pseudo-code in the Supp. \ref{supp:imp}.

\subsection{Dilated Phasor Volume \label{sec:volume}}
To preserve high frequencies, spatial embedders require discretizing the volume at a very fined level.
By contrast, frequency modeling manages to query specific frequencies without resorting to large volume size and memory   \citep{tancik2020fourfeat}. 
To this end, we model the sparse Fourier representation (with many entries masked) of the scene and augment it with neural networks to achieve full spectrum reconstruction. 
We use a small 3D optimizable phasor volume $\rmP^{\rw} \in \mathbb{C}^{k\times \rn \times  \rn \times \rd}$ to parameterize a full Fourier volume $\mathbb{C}^{k\times \rn \times  \rn \times \rn}$ that with certain frequency masked. In other words, we dilate $\rmP^\rw$ into $\rmP$. To illustrate, 
we set the frequencies on $\rvu,\rvv$ to be linear-increased, \emph{i.e.}, $\rvu = \rvv = \{\floor{-\rn/2}, \floor{-\rn/2}+1, \cdots, \floor{\rn/2}\}$, 
and set the frequencies on $\rvw$ to be exponential increased $\rvw=\{0,1,\cdots, 2^{\rd-2}\}$, following \cite{mildenhall2020nerf}. In this way, the frequency $\rmK = \textrm{meshgrid}(\rvu,\rvv,\rvw) \in \mathbb{R}^{\rn \times  \rn \times \rd\times 3}$. To avoid biasing axis $\rvw$, we use two additional volumes to parameterize $\rmP$, \emph{i.e}, $\rmP=\text{dilate}[\rmP^\rvu]+\text{dilate}[\rmP^\rvv]+\text{dilate}[\rmP^\rvw]$, with the superscript denoting the dilated dimension.
In Fig. \ref{algo:PREF}(b), we showcase the dilated $\rmP^\rw$ and $\rmP^\rw$ in yellow color.
As Fourier transform is linear, we have
\begin{equation}
    \rvf(\rvx) = \mathcal{T}({\rmP}) = \mathcal{T}(\rmP^\rvu) + \mathcal{T}(\rmP^\rvv) + \mathcal{T}(\rmP^{\rvw}) = \rmP  \textrm{exp}(-j2\pi\rmK\rvx). 
\end{equation}
This way, we can encode the coordinate with dense multi-dimensional spatial frequencies, mitigating the need for deep neural networks and longer training time to achieve full spectrum reconstruction. If only a few on-axis dilated frequencies comprise the support, our PREF degenerates to the Positional Encoding \citep{mildenhall2020nerf}, where the first linear layer can be regarded as $\rmP$. Therefore, our method is a scalable solution to trade off memory and speed. $\rmK$ serves as hyper-parameters. In the following, we discuss how to accelerate $\mathcal{T}$ and drop the superscript of $\rmP^\rw$ for clarity.

\subsection{Approximated Inverse Fourier Transform \label{sec:aft}}
Directly applying Fourier mapping \citep{tancik2020fourfeat} from such a large phasor volume is prohibitively expensive. To make the computation tractable, we can approximate it via first computing and storing a transformed embedding grid, from which embedding at arbitrary locations can be efficiently locally interpolated. 
Since we assume our underlying embedding field is smooth, the approximation error is modest when the grid size is sufficiently large. 
To illustrate, assuming we transform the \textit{full phasor volume} $\rmP\in\mathbb{C}^{\rk\times \rn\times\rn\times\rn}$, we can first transform and store it as a spatial grid $\mathbb{R}^{\rk\times \rn\times\rn\times\rn}$. This process equals the inverse discrete Fourier Transform (IDFT) that can be efficiently solved by well-established fast Fourier Transform (FFT) methods, e.g., the Cooley–Tukey algorithm \citep{cooley1965algorithm}. In each forward iteration, the inverse FFT is performed once for all coordinates so that $\mathcal{T}$ can be efficient. Fig. \ref{fig:overview} presents the conceptual idea. Notice this na\"ive solution has redundant computations, as this \textit{full sparse phasor volume} contains many zero entries.

To tailor our dilated phasor volume, we employ both IFFT and Fourier mapping to achieve high accuracy and low complexity.
Specifically, we perform 2D IFFT along the non-dilated axes $\rvu, \rvv$ to obtain an intermediate volume, a partially transformed volume, \emph{i.e.},
\begin{equation}
\rmF[\rx^*,\ry^*,:]
\footnote{We slightly abuse the matrix indexing, where $\rx^*$ and $\ry^*$ should be mapped to integers.}
= \sum_{\ri=0}^{M}\sum_{\rj=0}^{N}\re^{j2\pi(\rx^*\ru_\ri+\ry^*\rv_\rj)} \rmP[{\ri,\rj,:}] = \text{iFFT2D}(\rmP[\ri,\rj,:]) 
\end{equation}
Now, each $\rmF[\rx^*,\ry^*,:]$ only stores Fourier coefficients along the dimension $\rz$. In the subsequent, linear interpolation and a per-sample hermitian-symmetric Fourier mapping along $\rz$ are conducted:
\begin{equation}
\begin{split}
    \rvf(\rx, \ry, \rz)
    \approx 
    \sum_{k=0}^{d} 
    \underbrace{
    \re^{j2\pi \rw_k\rz}  \widetilde{\rmF}[\rx,\ry,\rk] + \re^{-j2\pi \rw_k\rz} 
    \widetilde{\rmF}^*[\rx,\ry,\rk]
    }_\text{Hermitian-symmetric Fourier mapping} \\
    = \sum_{k=0}^{d} 2\text{Re}(\widetilde{\rmF})[\rx,\ry,\rk] \sin{(2\pi\rw_k\rz)} - 2\text{Im}(\widetilde{\rmF})[\rx,\ry,\rk]\cos{(2\pi\rw_k\rz)}
    \label{eq:ifft}
\end{split}
\end{equation}
where $\widetilde{\rmF}$ is a continuous phasor field bilateral-interpolated from the phasor grid $\rmF$. 
We force the transformed feature to be real quantity by assuming the phasor field to be hermitian-symmetric, \emph{i.e.}, $\widetilde{\rmF}[-\rx, -\ry,-\rk]$ is $\widetilde{\rmF^*}[\rx, \ry,\rk]$, the conjugate of $\widetilde{\rmF}[\rx, \ry,\rk]$. 
Note that the length $\rd$ in the dilated dimension is generally small. Therefore, per-sample Fourier mapping is very efficient, significantly reducing the training cost. 

\subsection{Volume Regularization\label{sec:reg}}

\begin{figure}
\vspace{-20pt}
    \centering
    \begin{minipage}{0.44\textwidth}
        \centering
        \includegraphics[width=0.68\textwidth]{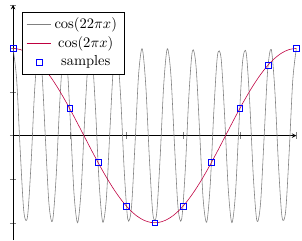} 
        \caption{Illustration on overfitting.}
        \label{fig:example}
    \end{minipage}
    \begin{minipage}{0.55\textwidth}
        \centering
        \includegraphics[width=0.84\textwidth]{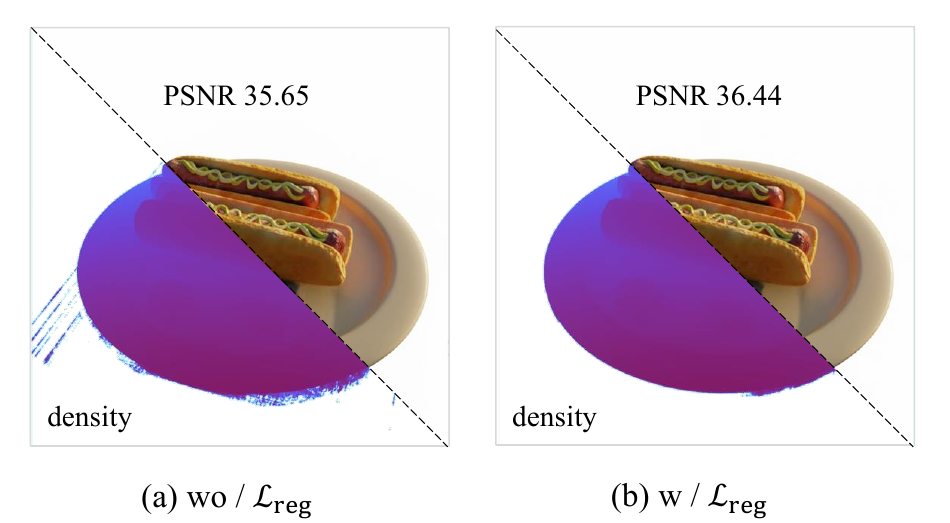}
        \caption{Effect of volume regularization.}
        \label{fig:tv}
    \end{minipage}
\end{figure}

In many reconstruction tasks, we confront a dilemma between overfitting and generalization due to limited supervision. Overfitting can occur, especially when the frequency bandwidth exceeds the minimal sampling rate of the observations (training samples). For example, training samples at a 10 Hz sampling rate from a continuous function $\cos(2\pi x)$ can also be fitted exactly by $\cos(22 \pi x)$, where the latter can be treated as overfitting in the lack of prior knowledge. See Fig. \ref{fig:example} for illustration. Therefore, when increasing the sampling rate (\emph{i.e.}, training samples) is not applicable, it is desirable to regularize the representations. Despite plenty of regularizers existing, for example, $L2$ norm (weight decay) for MLP and TV regularizer for discrete grid embedders, they are not directly applicable to frequency encoders. 
The Fourier coefficient should fall quickly to zero to smooth the embedding fields as frequency increases. To this end, we propose a simple and effective Parsvel regularizer:

\begin{equation}
    \begin{aligned}
            \mathcal{L}_\text{reg} 
            & = ||2\pi\ru\rmP||_2 + ||2\pi\rv\rmP||_2 + ||2\pi\rw\rmP||_2 \\
            & = \sqrt{\sum_{\ri,\rj,\rk}||2\pi \ru_\ri \rmP[\ri,\rj,\rk] ||^2} + \sqrt{\sum_{\ri,\rj,\rk}||2\pi \rv_\rj \rmP[\ri,\rj,\rk] ||^2} + \sqrt{\sum_{\ri,\rj,\rk}||2\pi \rw_\rk \rmP [\ri,\rj,\rk]||^2}
    \end{aligned}
\end{equation}

This regularizer resembles $L_2$ norm used in MLP, except that each entry is multiplied by its spatial frequency. 
the Parseval regularizer equals the anisotropic TV regularizer in the spatial domain. A proof based on Parseval's theorem is included in the Supp. \ref{supp:reg}. Implementation-wise, the regularizer can be integrated into modern optimizers without extra cost.


%% file: iclr2023/section/4_experiment.tex
\section{Experiments}

\setlength{\tabcolsep}{3pt}
\begin{table}[t]
\vspace{-15pt}
\caption{Fast radiance fields reconstruction on the NeRF-synthetic dataset \citep{mildenhall2020nerf}.}
\footnotesize
\centering
{%
\begin{tabular}{l|cc|cc|cc}
\toprule
Method & BatchSize & Steps & Time $\downarrow$  & Size(MB)$\downarrow$ & PSNR$\uparrow$ & SSIM$\uparrow$  \\
\hline
    SRN~\citep{sitzmann2019scene}          &  - & -       & $>$10h     & - &22.26 & 0.846  \\
    NeRF~\citep{mildenhall2020nerf}        & 4096 & 300k  & $\sim$35h  &  5.0    & 31.01 & 0.947 \\
    SNeRG~\citep{hedman2021baking}         & 8192 & 250k  & $\sim$15h      & 1771.5& 30.38 & 0.950      \\
    \hline
    NSVF~\citep{liu2020neural}             & 8192 & 150k  & $>$48h &  -    & 31.75 & 0.950\\
    PlenOctrees$\dag$~\citep{yu2021plenoctrees}  & 1024 & 200k  & $\sim$15h      & 1976.3& 31.71 & 0.958   \\
    Plenoxels$\dag$~\citep{yu2021plenoxels}      & 5000 & 128k  & 11.4m  & 778.1 & 31.71 & 0.958   \\
    DVGO~\citep{sun2021direct}             & {5000} & 30k   & 15.0m          & 612.1 & 31.95 & 0.957 \\
    NGP$\dag$~\citep{muller2022instant}         & 10k$\sim$85k   & 30k   & 4.6m & 87.3 & 32.63 & 0.921\\
    TensoRF~\citep{chen2022tensorf}        & 4096   & 30k   & 17.4m &71.8 & 33.14 & 0.963\\
\hline
Ours {\color{gray} PREF-128} & 4096 & 30k & $\sim$ 18m & 34.4 & 32.08 & 0.952  \\
Ours {\color{gray} PREF-64} & 4096 & 30k & {$\sim$18m} & 9.84 & 31.18 & {0.950}  \\
Ours {\color{gray} PREF-32} & 4096 & 30k & {$\sim$18m} & 2.28 & 29.83 & {0.940} \ \\
\hline
\end{tabular}
} \par
{ $\dag$ denotes using customized Cuda kernels.}
\vspace{-5pt}
\label{table:rendering_score}  
\end{table}
\setlength{\tabcolsep}{1.4pt}

\subsection{Neural Radiance Fields Reconstruction}

For the neural radiance field, we focus on rendering novel views from a set of images with known camera poses. Each RGB value in each pixel corresponds to a  ray cast from the image plane. We adopt the volume rendering model used by \cite{mildenhall2020nerf}:

\begin{equation}
    \hat{C}(r) = \sum_{i}^N T_i(1-\mathrm{exp}(-\sigma_i \delta_i))c_i, \text{where  } T_i = \mathrm{exp}(-\sum_{j=1}^{i-1}\sigma_j\delta_j)
\end{equation}

where $\sigma_i$ and $c_i$ are corresponding density and color at location $\mathrm{x}_i$, $\delta_i$ is the interval between adjacent samples. Then we optimize the rendered color with the ground truth color with the following loss:
\begin{equation}
    \mathcal{L} = \frac{1}{M}\sum_{i=0}^{M} ||C(r) - \hat{C}(r)||^2 + \lambda \mathcal{L}_\text{Reg}.
\end{equation}
\textbf{PREF Setting.} We describe how PREF models the density $\sigma$ and the radiance $c$. We use three phasor volumes of $256 \times 256 \times 1$ width $16\rd$ feature length augmented with a two-layer MLP with hidden dimension $64\rd$. We then use Softplus to map the raw output to the positive-valued density. For the view-dependent radiance branch, we use a volume of $256 \times 256 \times 1$ with $32\rd$ feature length followed by a linear layer to output $27\rd$ feature embedding. The feature embedding therefore has an approximated Nyquist frequency of 128. To render 5D view-dependent radiance, we concatenate the result with the positional encoded view directions and feed them into a 2-layer MLP with $128\rd$ hidden dimension and a linear layer to map the feature to color with Sigmoid activation. All linear layers except the final use ReLU activation, following common practices.

\textbf{Rendering.} To compare with SOTAs \citep{sun2021direct, chen2020tensor}, we train each scene using $30\rk$ iterations with a batch size of $4096$ rays. We adopt a progressive training scheme \citep{hertz2021sape}: from the resolution of $128$ to $256$. Specifically, we gradually unlock the higher frequencies at the training step $[2000, 3000, 4000, 5500, 7000]$. Accordingly, the number of samples per ray progressively increases from about 384 to about 1024. This allows us to achieve faster and more stable optimization by first covering the lower frequencies and later high-frequency details. During training, we maintain an alpha mask to skip empty spaces to avoid unnecessary evaluations. We also skip evaluating rgb value as if that sample weight less than a small threshold $1e^{-4}$.

\textbf{Optimization.} As mentioned in the paper, our PREF uses the Parsvel regularizer $\mathcal{L}_\text{Reg}$ on the density phasor volume to avoid overfitting. Our objective is then set to $\mathcal{L} = \mathcal{L}_\text{RGB} + \lambda \mathcal{L}_\text{Reg}$ with $\lambda=1e^{-2}$. Without regularization, PREF may overfit specific frequencies, as shown in Fig. \ref{fig:tv}. On the NeRF synthetic dataset, PREF converges on average $18$ minutes with $30\rk$ iterations on a single RTX 3090, with a learning rate gradually decaying from $0.02$ to $0.002$ during the training. The Adam optimizer is used with $\beta_1=0.9$ and $\beta_2=0.99$ by default. \par
\begin{figure}[t]
    \centering
    \includegraphics[width=0.95\textwidth]{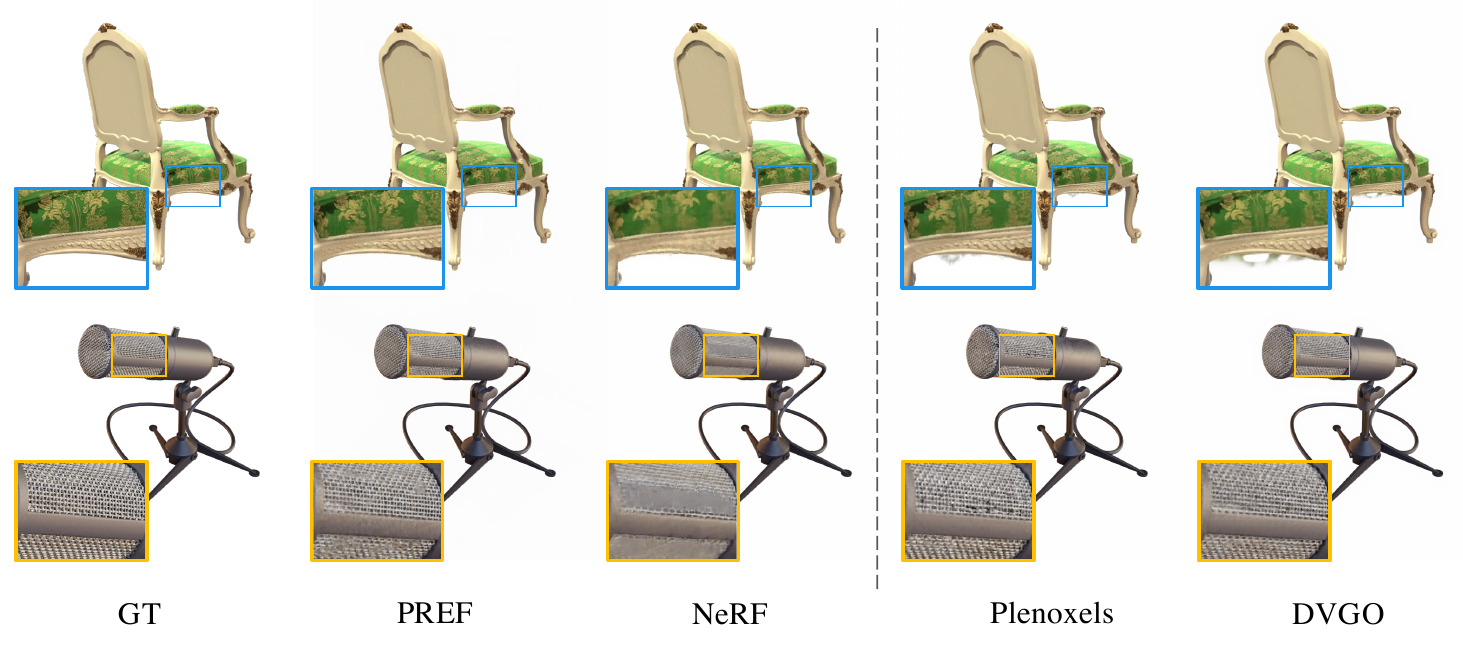}
    \caption{Qualitative comparisons. We compare to NeRF \citep{mildenhall2020nerf} which encode coordinates with fewer frequencies and larger MLP. We show by covering border spectra, 
    PREF not only reduces the costly MLP but also significantly improves the visual quality.
    We also compare frequency modelings to dense volume methods \citep{yu2021plenoxels, sun2021direct}, which tend to produce outliers in local regions. }
    \label{fig:nerf_vis}
\end{figure}

\textbf{Comparisons.} We use the blender dataset \citep{mildenhall2020nerf} for training and evaluation since plenty of public benchmarks with speed are available. We report the quantitative and qualitative comparisons in Tab. \ref{table:rendering_score}, Fig. \ref{fig:nerf_vis} respectively. We show that our PREF outperforms its dense volume counterparts and many proceeding designs, whereas NGP \citep{muller2022instant} and TensoRF \citep{chen2022tensorf} still present the best image quality.
TensoRF, the concurrent work, which factorizes the 3D volume with Tensor decomposition \citep{chen2020tensor}, seems to show a more effective inductive bias to exploit the correlation within scenes. To match their results, one promising attempt is to combine PREF and Tensor decomposition by reusing the separability of Fourier transform. NGP is also excellent to recover high frequency locals by distributing coordinates to different buckets with customized hash functions. Despite this, the choice of the hash function can be sensitive to the performance and the hashed features may not be directly applicable in generalization tasks. 


\textbf{Ablations}. We also include results of using smaller phasor volumes that have $128$ and $64$ resolution, respectively. We denote them as PREF-64 and PREF-32 according to their Nyquist frequency. We find that PREF with a higher volume size does not yield a better trade-off between memory and speed. Following \citep{yu2021plenoxels,sun2021direct, chen2022tensorf}, we can also model the density field with phasor volume without any MLP, denoted as PREF-vanilla. In this case, the density field is the summation of sinusoids, similar to FOM \citep{jansen2010fourier}. We tune phasor volumes with size $256\times256\times 10$ and achieve a similar PSNR $31.79$, comparable to that augmented with MLP. We advocate the lateral design since it provides a unified form for both density and radiance. Imposing regularization on the density field slightly improves the average PSNR from $31.62$ to $32.08$. 
Without regularization, the density field tends to be cloudy and hinders the effectiveness of the alpha mask, which roughly doubles the training time. 

\begin{table}[t]
\vspace{-15pt}
\caption{Quantitative results of SDF regression.}
    \centering
    \footnotesize
    \begin{tabular}{c|cc|lcclcc}
        \toprule
        \multirow{2}{*}{Method}
        & \multirow{2}{*}{Size(MB)}
        & \multirow{2}{*}{Time}
        &
        & \multicolumn{2}{c}{Armadillo}
        &
        & \multicolumn{2}{c}{Gargoyle} \\
        \cline{4-6} \cline{8-9}
        \multirow{2}{*}{} & \multirow{2}{*}{}&\multirow{2}{*}{} &
         &IOU$\uparrow$ & Chamfer$-L_1$$\downarrow$ && IOU$\uparrow$ & Chamfer$-L_1$ $\downarrow$  \\
        \hline
        DV \citep{sun2021direct} & 128.0 & $\sim$21m && 98.81  & 5.67e-6 && 97.99 & 1.19e-5\\
        NGP \citep{muller2022instant} & 46.7 & $\sim$17m  && 99.34  & 5.54e-6 && 99.42 & 1.03e-5\\
        \hline
        PE \citep{tancik2020fourfeat}&  6.0& $\sim$26m && 96.65  & 1.21e-5 &&  80.46 & 1.03e-4 \\
        BACON \citep{lindell2022bacon} & 2.1 & $\sim$1.5h  &&  98.33 & 2.20e-6 && 98.76 & 3.65e-5\\
        Ours &  36.0 & $\sim$21m && 99.02 & 5.57e-6 && 99.05 & 1.06e-5 \\
        \bottomrule
    \end{tabular}
    \vspace*{-5pt}
    \label{tab:sdf}
\end{table}

\subsection{3D Shape Representation}

Next, we conduct the task of signed distance field (SDF) regression to evaluate its fitting capacity. 
SDFs describe the shape in terms of a function $f(\rx)$ as the signed Euclidean distance from the point $\rx$ to the surface.
Our goal is to recover a continuous SDF $f(\rx)$ given a set of sampled $f(\rx^*)$.

\textbf{Data preparation and training.} We adopt two widely used models: gargoyle (50k vertices) and armadillo (49k vertices). We scale the model within a bounding box of $[-1,1]$ and samples $N=2^{18}$ points for each training epoch: $4/8N$ points on the surface, $3/8N$ points around the surface by adding Gaussian noise to the surface point with $scale=0.01$, and the last $1/8N$ points uniformly sampled within the bounding box. In each training epoch, we sample a batch size of $N=2^{18}$ to regress the SDF values. The MAPE loss is used for error back-propagation. 
To optimize the networks, we use the Adam optimizer, with $\beta_1=0.9$, $\beta_2=0.99$ and $\epsilon=1e^{-5}$. We use an initial learning rate $1e^{-4}$ and reduce the learning rate to $1e^{-5}$ at the 10th epoch. We adopt a batch size of $N/100$ to optimize all baselines whereas for our method $20$ epochs.

\textbf{Metrics.} We report the IOU of the ground truth mesh and the regressed signed distance field by discretizing them into two $128^3$ volumes. We also report the Chamfer distance metric by sampling 30k surface points from the extracted mesh with Marching Cube \citep{lorensen1987marching}. 

\textbf{Baseline implementation.} 
For our PREF, we use three $16\times 128\times 128\times 6$ complex-valued volumes to nonlinearly transform the spatial coordinates to a $16\rd$ feature embedding.
For the embedding-based baselines, we use our implementation of the dense volume technique. It contains $16 \times 128 \times128 \times 128$ learnable parameters that transform the input coordinates to their feature embedding at a length $16$ by trilinear interpolation.
We also implement NGP \citep{muller2022instant}, where they maintain a multi-level hash function to transform the spatial coordinate into feature embedding. We use a 16 num-of-level hash function with dimension 2. Consequently, the output feature embedding is of length 32. All the embedding-based baselines and our PREF adopt the same MLP structure that consists of 3 layers that progressively map the input embedding to the 64 hidden dimension features and finally to a scalar, with ReLU as the intermediate activation.
For Position Encoding \citep{mildenhall2020nerf}, we use 6 log-sampled frequencies that roughly achieve 128 resolution.
To compensate for such pure implicit models,  we use a wider and deeper MLP configuration: 8 linear layers with 512 hidden dimensions. For BACON \citep{lindell2022bacon}, we direct follow their training configuration to obtain the results. We conduct the experiments on a single RTX 2080Ti.

\textbf{Comparisons.} Tab. \ref{tab:sdf} lists model size and performance of the baselines vs. PREF. Our method manages to match the SOTA NGP \citep{muller2022instant} with compact model size, and outperform its spatial counterpart dense volume and other frequency-based proceedings. We owe the improvement of PREF to its globally continuous nature that allows for preserving details. NGP also successfully capture high-frequency details without distortion, potentially owing to their multi-scale coordinate inputs that capture global information. PREF only requires a single coordinate input because the multi-scale signals have been encoded in the frequency space.

\begin{wrapfigure}[7]{r}{0.5\textwidth}
\vspace{-10pt}
\begin{center}
\includegraphics[width=0.5\textwidth]{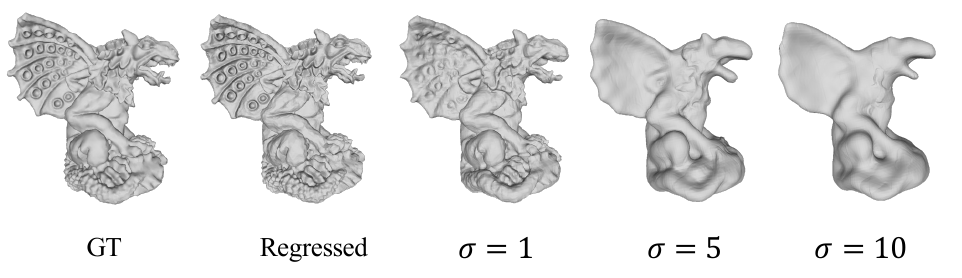}
\end{center}
\vspace{-8pt}
\caption{Neural field smoothing.}
\label{fig:sdf_smoothing}
\end{wrapfigure}

\textbf{Level of detail filtering}.  In line with frequency representations  \citep{lindell2022bacon,barron2021mip}, 
our PREF also allows implicit surface editing \citep{yang2021geometry} such as level of detail filtering. 
But different from them, our method does not need multi-scale supervision, since we have assigned a major capability of the network to the interpretable phasor embedding fields. Therefore, we can manipulate the trained phasor embedding directly such as multiplying a kernel, which will result in the convolution of transformed fields and that kernel. 
For example, point-wise multiplying with Gauss kernels of different variations $\sigma$ yields neural fields of different level-of-details, as Fig. \ref{fig:sdf_smoothing} shows.  Details are included in Supp \ref{supp:ldf}.
Such multi-scale representation can benefit many
computer graphics tasks like texture removal. Moreover, other predefined or leaned kernels may be further used to deform the neural fields in different styles. To some extent, this experiment also explains why PREF is robust to recovering high frequencies, since a slight change in phasor magnitude will result in global details recovery or suppression.

\begin{figure}[t]
\vspace{-15pt}
    \centering
    \includegraphics[width=.85\textwidth]{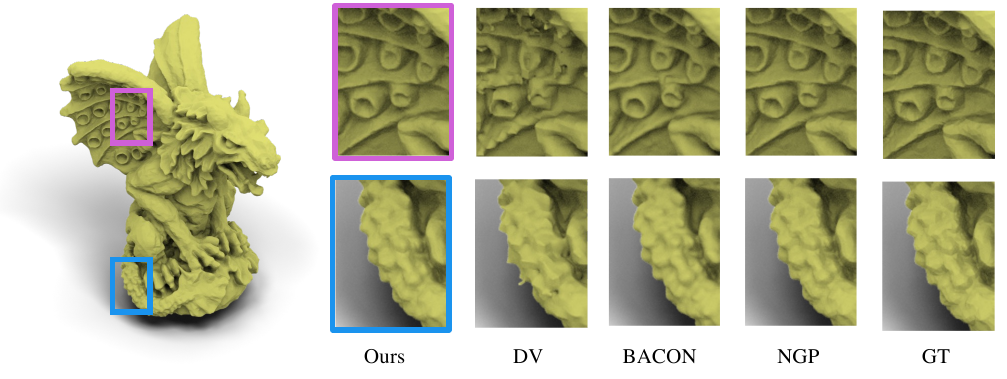}
    \caption{Qualitative visualizations of regressed SDF. The left image shows our regressed result. On the right, we compare our method to a dense volume encoder, BACON \citep{lindell2022bacon} and NGP \citep{muller2022instant}.
    \vspace{-10pt}
    }
    \label{fig:sdf}
\end{figure}

\subsection{understanding the inductive bias of frequency representation}
\begin{figure}[t]
\centering
\begin{minipage}{0.52\textwidth}
\centering
\includegraphics[width=1\textwidth]{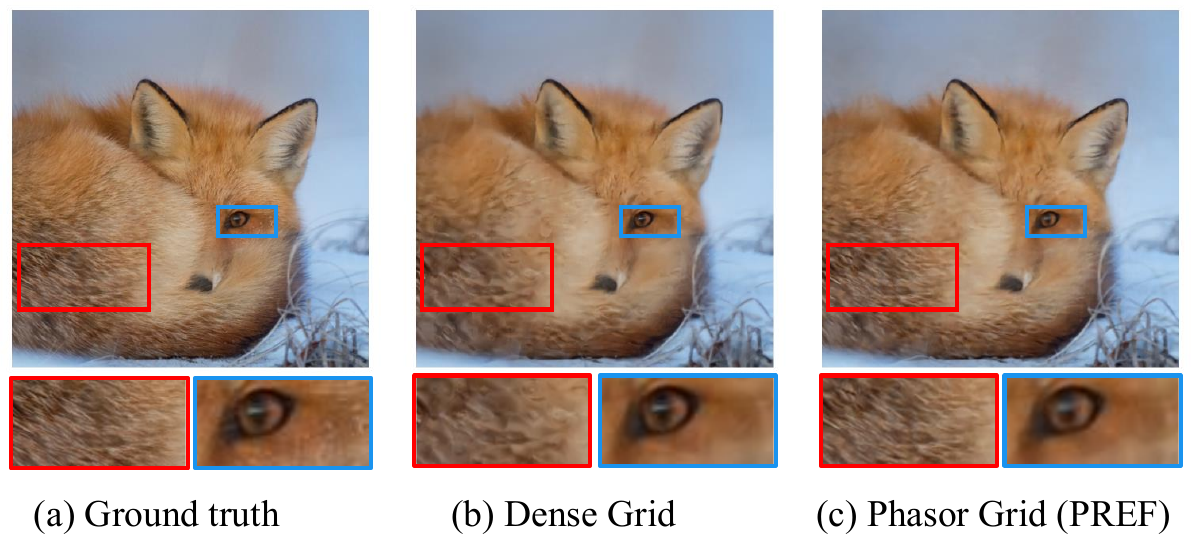} 
\captionof{figure}{\small Image Regression (from Natural dataset).}
\label{fig:image_regression}
\end{minipage}
\hfill
\begin{minipage}{0.45\textwidth}
\centering
\captionof{table}{ Image generalization (PSNR). We compare to PE \citep{mildenhall2020nerf}, SIREN \citep{sitzmann2019siren} and NGP \citep{muller2022instant}.}
\vspace*{-3pt}
\tiny
\begin{tabular}{cccc}
        \toprule
        &Memory (KB)& Natural & Text \\
        \hline
        Dense Grid & $838$ & $23.627\pm  4.085$   &  $27.561 \pm 2.400$  \\
        NGP & $696$ &   $24.137 \pm  3.509$    & $27.641 \pm 1.852$ \\
        \hline
        PE & $1318$ &   $22.294 \pm 3.245$   &  $26.852 \pm 1.860$    \\
        SIREN & $777$ &   $ 22.394 \pm 2.347$   &  $25.114 \pm 6.680$    \\
        Ours &  $545$ &  $24.113 \pm 3.464$    & $28.329 \pm 2.373$ \\
        \bottomrule
    \end{tabular}
    \label{tab:image_generations}
\end{minipage}
\vspace{-5pt}
\end{figure}

Despite there are theoretical analyses of the inductive bias of Fourier encoding from the perspective of NTK \citep{tancik2020fourfeat} or dictionary learning \citep{yuce2022structured}, we still get less insight into how frequency representation compares to dense grid representation.
Theoretically, PREF with linear frequency distribution has the same expressiveness as a dense volume with the same volume size. Despite this, we notice that their expertise is different: PREF excels at recovering textures while dense grid excels at recovering high-fidelity local regions. For example, Fig. \ref{fig:image_regression} shows that PREF better captures the fur while Dense Grid better captures the eye. Despite both two encoders having the same Nyquist frequency and volume size, PREF achieves slightly better PSNR (31.86) than Dense Grid (31.02) due to the large portion of fur. We owe this difference to the optimization behaviors: PREF 
synthesize a position with periodic sinusoidal
while Dense Grid synthesise it with local parameters. 
For further validation and comparisons to more advanced approaches, we conduct the benchmark image painting task \citep{huang2021textrm} which aims to regress the whole image only given 25\% pixels. Results are shown in Tab. \ref{tab:image_generations}. Implementation detail is in the supplementary.

%% file: iclr2023/section/6_conclusion.tex
\section{Limitations}
Speed-wise, PREF incurs additional training time caused by FFT but not in inference as we can pre-store the transformed features. Consequently, it is slightly slower in training but comparably fast in inference to SOTA. As opposed to a global FFT that transforms the entire phasor volume, a local window-based FFT, also known as short-time FFT, may leverage both frequency sparsity and spatial sparsity to further save memory and potentially accelerate training. Quality-wise, to maintain tiny volumes, PREF dilates along the reduced dimensions to cover the full spectra. As a result, it may introduce axial-align bias, similar to \cite{mildenhall2020nerf}. To mitigate the problem, one may use NuFFT (Non-uniform Fast Fourier Transform) \citep{fessler03} to achieve non-uniform frequency sampling \citep{tancik2020fourfeat} as well as fast Fourier transform.

\section{Conclusion}
The main motivation behind implicit neural representations is to search for \textit{fast} algorithms to compute \textit{compact} representation of functions and datasets. In this work,  we extend the family of frequency representations by presenting PREF, a novel approach that combines neural networks and Fourier serials for compact representations. 
PREF remains \textit{fast} by an approximated fast Fourier transform and retains \textit{compact} by a tailored dilated phasor volume. PREF produces high-quality images, shapes, and radiance fields from given limited data and shows comparative results to the SOTAs. We also show our representation is more robust, potentially owing to its smooth and periodic properties that explore the correlations of data. In the future, we may apply our representation to efficient 3D-aware generative tasks, which may provide alternative frequency-based latent space for optimizations or controls.

%% file: iclr2023/section/7_appendix.tex
\begin{center}
\Large \textbf{Supplementary}
\end{center}
\vspace{10pt}

\section{Parsvel Regularization \label{supp:reg}}
\begin{theorem}[Differentiation Theorem]
\label{th:diff}
Let $\rmF(\ru,\rv)$ be an absolutely continuous differentiable function, and $\rf(\rx,\ry,\rz)$ be its inverse Fourier transform, we have 
\begin{equation}
\frac{\partial \rf^n(\rx,\ry, \rz)}{\partial \rx^n} = \mathcal{T}{((\rj 2\pi \ru)^n \rmF(\ru,\rv, \rw))} \label{eq:differentiation}
\end{equation}
\end{theorem}

\begin{theorem}[Parsvel Theorem]
\label{th:par}
Let $\rmF(\ru,\rv)$ be absolutely continuous differentiable function, and $\rf(\rx,\ry,\rz)$ be its inverse Fourier transform, we have 
\begin{equation}
\iiint ||\rf(\rx,\ry,\rz)||^2 \rd\rx\rd\ry\rd\rz = 
\sum_{\ru,\rv,\rw}
||\rmF(\ru,\rv, \rw)||^2. \label{eq:parseval}    
\end{equation}
\end{theorem}

\begin{lemma}
Let $\rf(\rx,\ry,\rz)$ be integrable, and $\rmP(\ru,\rv, \rw)$ be its Fourier transform. 
The anisotropic TV loss of $\rf(\rx,\ry,\rz)$ can be represented by $||2\pi\ru\rmP||_2 + ||2\pi\rv\rmP||_2 + ||2\pi\rw\rmP||_2$
\end{lemma}

\textit{\textbf{Proof:}}
Recall the TV loss can be computed as $||\nabla_x \rf(\rx,\ry,\rz)||_2 + ||\nabla_y \rf(\rx,\ry,\rz)||_2 + ||\nabla_z \rf(\rx,\ry,\rz)||_2$. Since $\rf(\rx,\ry,\rz)$ and $\rmP(\ru,\rv,\rw)$ are Fourier pairs, we have Fourier transform preserves the energy of original quantity based on Parseval's theorem (theorem \ref{th:par}), \emph{i.e.},
\begin{equation}
    \iiint ||\rf(\rx,\ry,\rz)||^2 \rd\rx\rd\ry\rd\rz = 
    \sum_{\ri,\rj,\rk} ||\rmP[\ri,\rj,\rk]||^2. 
\end{equation}
According to theorem \ref{th:diff}, $\nabla_x \rf(\rx, \ry, \rz)$ and $\rj2\pi \ru\rmP(\ru,\rv, \rw)$ are also Fourier pairs. The integration derivative along axis $x$ is defined as,
\begin{equation}
    \iiint||\nabla_x \rf(\rx,\ry,\rz)||^2\rd\rx\rd\ry\rd\rz = 
    \sum_{\ri,\rj,\rk}||\rj2\pi \ru_\ri\rmP[\ri,\rj,\rk]||^2
\end{equation}
By taking square root on both sides, we have $||\nabla_x\rf(\rx,\ry,\rz)||_2 = ||2\pi  \ru\rmP(\ru,\rv,\rw)||_2$. And $||\nabla_y\rf(\rx,\ry,\rz)||_2 = ||2\pi \rv\rmP(\ru,\rv,\rw)||_2$, $||\nabla_z\rf(\rx,\ry,\rz)||_2 = ||2\pi \rw\rmP(\ru,\rv,\rw)||_2$ can be derived similarly. 

\section{Phasorial Embedding Fields Implementation Details \label{supp:imp}}
\textbf{Dilated Phasor Volume.}
Recall that PREF is a continuous embedding field corresponding to a multi-channel multi-dimensional square Fourier volume. We elaborate on implementation details. We model the underlying sparse spectra of the scene as feature embedding, and augment it with neural network to achieve full spectrum recovery. Let $\mathbf{P}[u,v,w]$ be a 3D phasor volume representing an embedding field $f(x,y,z)$. Note that the $\rmP[u,v,w]$ is Hermitian symmetric when the $f(x,y,z)$ is a real-valued feature embedding,
\emph{i.e.}, $\rmP[u,v,w] = \rmP^{*}[-u,-v,-w]$ (i.e., its complex conjugate). 
Further, based on the observation that natural signals are generally band-limited, we model their corresponding fields with sparse band-limited phasor volumes $\mathbf{P}[u,v,w]$ where we are able to partially mask out some entries and factor the volume along respective dimensions, as shown in Fig \ref{fig:overview}. Thus we arrange the full spectrum into tri-thin embeddings by reusing the linearity of the Fourier Transform, $f(x,y,z) = \mathcal{T}(\rmP_u) + \mathcal{T}(\rmP_v) + \mathcal{T}(\rmP_w)$.
\begin{figure}[h]
    \centering
    \includegraphics[width=0.8\textwidth]{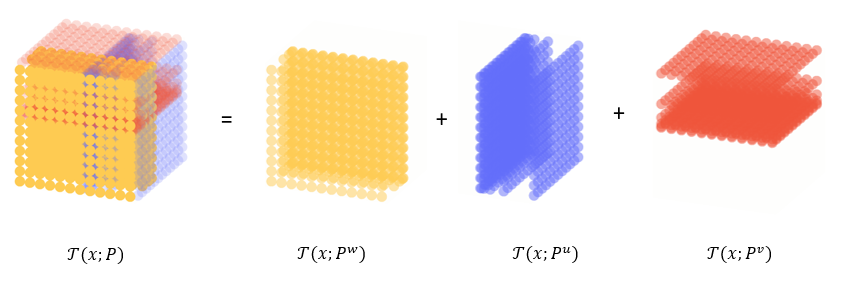}
    \caption{Illustration on arranging a full sparse volume into three compact volumes. Let the phasor volume with zero frequency centered.We selectively mask out some entries of the phasor volume as zero, then approximate the spatial feature embedding of a large phasor volume in terms of the sum of the embedding from three smaller ones, each of which dilates along an axis. Despite there are some overlap entries, \emph{e.g.}, the center entry appears in all three tiny volumes, it will not affect the performance as it is an over-parameterization. The three compact volumes can be reparameterized back to the original single full volume for conducting Fourier multiplication, such as level-of-detail filtering. The complete transformation from phasor volume to spatial feature embeddings is shown in the PyTorch pseudo-code in Algorithm \ref{algo:PREF}.}
    \label{fig:implementation}
\end{figure}
\par
\textbf{IFT Implementation.}
Recall that for a 3D phasor volume, we approximate $\mathcal{T}$ via sub-procedures of applying 2D Fast Fourier Transforms (FFTs) and 1D Fourier mapping to achieve high efficiency. Therefore, given a batch of spatial coordinates, our PREF representation transforms them into a batch of feature embeddings in parallel where PREF can serve as a plug-and-play module. Such a module can be applied to many existing implicit neural representations to conduct task-specific neural field reconstructions. We present a sketchy PyTorch pseudo-code in Algorithm \ref{algo:PREF}.

\begin{algorithm}[H]\small
	\caption{PREF Encoder in a PyTorch-like style.}
	\definecolor{codeblue}{rgb}{0.25,0.5,0.5}
	\lstset{
		backgroundcolor=\color{white},
		basicstyle=\fontsize{7.2pt}{7.2pt}\ttfamily\selectfont,
		columns=fullflexible,
		breaklines=true,
		captionpos=b,
		commentstyle=\fontsize{8pt}{8pt}\color{codeblue},
		keywordstyle=\fontsize{8.0pt}{8.0pt},
	}
\begin{lstlisting}[language=python]
import torch
import torch.nn as nn

class PREF(nn.Module):
    def _init_(self, res, d, ks):
        """
        res: resolution size
        d: reduced dim size
        ks: output kernel size
        """
        Nx, Ny, Nz = res 
        # log sampling freq in reduced dimension
        self.freq = torch.tensor([0]+[2**i for i in torch.arange(d-1)]) 
        self.Pu = nn.Parameter(torch.zeros(1, ks, d, Ny, Nz).to(torch.complex)) 
        self.Pv = nn.Parameter(torch.zeros(1, ks, Nx, d, Nz).to(torch.complex)) 
        self.Pw = nn.Parameter(torch.zeros(1, ks, Nx, Ny, d).to(torch.complex)) 
    def forward(self, xyz): 
        # 2D Fast Fourier Transform
        Pu = torch.fft.ifftn(self.Pu, dim=(3,4)) 
        Pv = torch.fft.ifftn(self.Pv, dim=(2,4)) 
        Pw = torch.fft.ifftn(self.Pw, dim=(2,3)) 
        # 2D Linear interpolation
        xs, ys, zs = xyz.chunk(3, dim=-1) 
        Px = grid_sample_cmplx(Pu.transpose(3,3).flatten(1,2), torch.stack([zs, ys], dim=-1)[None]).reshape(Pu.shape[1], Pu.shape[2], -1) 
        Py = grid_sample_cmplx(Pv.transpose(2,3).flatten(1,2), torch.stack([zs, xs], dim=-1)[None]).reshape(Pv.shape[1], Pv.shape[3], -1) 
        Pz = grid_sample_cmplx(Pw.transpose(2,4).flatten(1,2), torch.stack([xs, ys], dim=-1)[None]).reshape(Pw.shape[1], Pw.shape[4], -1) 
        # 1D Fourier Mapping 
        fx = batch_NI(Px, xs, self.freq) 
        fy = batch_NI(Py, ys, self.freq) 
        fz = batch_NI(Pz, zs, self.freq) 
        # Summation
        return fx+fy+fz
\end{lstlisting}
\label{algo:PREF}
\end{algorithm}

\textbf{Phasor Volume Initialization.}
Our PREF approach can be alternatively viewed as a frequency space learning scheme for existing spatial coordinate-based MLPs. In our experiments, we found zero initialization works well for applications ranging from 2D image regression to 5D radiance field reconstruction while certain applications require more tailored initialization, e.g., geometric initialization in \citep{SAL}. This is because $\rmP(\mathrm{k})$ (with $\mathrm{k}$ being the frequency coordinate) needs to satisfy the unique constraints of $f(\mathrm{x})$. We thus initialize the phasor volume as follows: Let $f^{\circ}(\mathrm{x})$ be the initialization of $f(\mathrm{x})$. We have $\rmP^{\circ}(\mathrm{k})$ = $\mathcal{T}(f^{\circ}(\mathrm{x}))$, with $\mathcal{T}$ as the Fourier transform. We then transform $\rmP^{\circ}(\mathrm{k})$ via the inverse Fourier transform $\mathcal{T}$ (due to duality between $f(\mathrm{x})$ and $\rmP(\mathrm{k})$)  as the \textit{approximation} to $f^{\circ}(\mathrm{x})$. We found such a strategy enhances stability and efficiency. 

\textbf{Computation Time.} One of the key benefits of PREF is its efficiency. We conduct frequency-based neural field reconstruction by employing IFT, which is computationally low cost and at the same time effective. When the input batch is sufficiently large (e.g., $4096 \times 1024$ samples per batch in radiance field reconstruction), the per-sample numerical evaluation will dominate the computational cost. Since such per-sample evaluation can be efficiently implemented using matrix products, it is essentially equivalent to adding a tiny linear layer. The overall implementation makes PREF nearly as fast as the state-of-the-art, e.g., instant-NGP for NeRF. For example, on the Lego example, our PyTorch PREF produces the final result in 16 minutes on a single RTX3090, considerably faster than the original NeRF and comparable to the PyTorch implementation of NGP. We are in the process of implementing PREF on CUDA analogous, and hopefully, it may achieve comparable performance to the CUDA version of NGP. 

\renewcommand{\tabcolsep}{3pt}
\begin{table*}[t]
  \centering
  \scriptsize
  \begin{tabular}{llcccccccclc|lc}
    \toprule
     & & Chair & Drums & Ficus & Hotdog & Lego & Materials & Mic & Ship & & Mean & & Size (MB)$\downarrow$ \\ \cmidrule(){1-1} \cmidrule(){3-10} \cmidrule(){12-12} \cmidrule(){14-14}
    PlenOctrees \citep{yu2021plenoctrees} & & 34.66 & 25.37& 30.79 & 36.79 & 32.95 & 29.76 & 33.97 & 29.62 & & 31.71 && 1976.3 \\
    Plenoxels \citep{yu2021plenoxels} & & 33.98 & 25.35 & 31.83 & 36.43 & 34.10 & 29.14 & 33.26 & 29.62 & & 31.71 && 778.1  \\
    DVGO \citep{sun2021direct} & & 34.09 & 25.44 & 32.78 & 36.74 & 34.46 & 29.57 & 33.20 & 29.12 && 31.95 && 612.1 \\
     Ours  & & 34.95 & 25.00 & 33.08 & 36.44 & 35.27 & 29.33 & 33.25 & 29.23 & & 32.08 &&  34.4  \\
    \bottomrule
    \end{tabular}
    \caption{PSNR results on each scene from the Synthetic-NeRF dataset \citep{mildenhall2020nerf}. We show the comparisons of the dense volume variants with our PREF (frequency-based scheme).}
    \label{tab:break_down}
\end{table*}

\section{Image Regression \label{supp:img_reg}}
\textbf{Image regression and completion.} To evaluate PREF's robustness, we demonstrate PREF on image completion tasks. We use the commonly adopted setting \citep{huang2021textrm,fathony2020multiplicative}: given 25$\%$ pixels of an image, we set out to predict another $25\%$ pixels. We evaluate PREF vs. SOTA on two benchmark datasets - Nature and Text. Specifically, we compare PREF with a dense grid counterpart, and two state-of-the-art coordinate-based MLPs \citep{mildenhall2020nerf, sitzmann2019siren}. The dense grid uses a $8 \times100 \times 100$ resolution whereas PREF uses two $20\times 9$ grids that correspond to the highest frequency of $2^7=128$. The two embedding techniques above use the same MLP with three linear layers, 256 hidden dimensions, and ReLU activation. We use Positional Encoding (PE) which consists of a 5-layer MLP with $7$ frequencies encoding. We adopt SIREN from \citep{sitzmann2019siren} that uses a 4-layer MLP and sine activation. Detailed comparisons are listed in Tab. \ref{tab:image_generations}. 

\textbf{Optimization details}. All experiments use the same training configuration. Specifically, we adopt the Adam optimizer \citep{adam} with default parameters  ($\beta_1=0.9,\beta_2=0.999,\epsilon=1e^{-8}$), a learning rate of $1e^{-4}$. We use $L_1$ loss with $15k$ iterations to produce the final results.

\section{Level of Detail Filtering \label{supp:ldf}}
Recall that the continuous embedding field of PREF is synthesized from a phasor volume under various frequencies. Therefore, thanks to Fourier transforms, various tools such as convolution in the continuous embedding fields can be conveniently and efficiently implemented as multiplications. This is therefore a unique advantage of PREF compared with its spatial embedding alternatives \citep{chen2020tensor,sun2021direct,yu2021plenoxels, muller2022instant}. \par
Let $\mathcal{M}(;\theta)$ and $\rmP[u,v,w]$ be the optimized MLP and phasor volume, respectively. $\mathcal{T}$ represents the inverse Fourier Transform.
Recall that we obtain a reconstruction field by $\Phi(\mathrm{x}) = \mathcal{M}(\mathcal{T}( \rmP ;\mathrm{x});\theta)$. Modification to the original signal via convolution-based filtering can now be derived as:
\begin{equation}
    \Phi^*(\mathrm{x}) = \mathcal{M}(\mathcal{T}(\rmP \circ \mathrm{G} ;\mathrm{x});\theta)
\end{equation}
where $\circ$ denotes element-wise multiplication and $\mathrm{G} \in \mathbb{C}^{l\times N^3}$ is a filter. \par
Now, we explore how to manipulate $\Phi(\mathrm{x})$ via the optimized phasor volume $\rmP$ and kernel $\mathrm{G}$. For simplicity, we only the Gaussian filter $\mathrm{G}$ while more sophisticated filters can also be applied in the same. Assume 
\begin{equation}
    G(\mathrm{k}) = \mathrm{exp}(-\mathrm{k}^T\mathrm{k}\sigma^2),
    \label{eq:gaussian}
\end{equation}
where $\mathrm{k} = [u/N,v/N,w/N]$ and $\mathrm{G}$ covers the complete frequency span of $\rmP$; that is, we can scale the magnitude of phasor features frequency-wise. For example, by varying the Gaussian kernel size using $\sigma$, PREF can denoise the neural representation of the signal at different scales.